\documentclass[conference]{IEEEtran}

\usepackage[pdftex]{graphicx}

\usepackage[utf8]{inputenc}
\usepackage{longtable}
\usepackage{multirow}
\usepackage{booktabs}
\usepackage{hyperref}
\usepackage[dvipsnames]{xcolor}
\definecolor{mypink1}{rgb}{0.858, 0.188, 0.478}

\usepackage{amsmath}
\usepackage{bbm}

\begin{document}

\title{G-SimCLR: Self-Supervised Contrastive Learning with Guided Projection via Pseudo Labelling}

\author{\IEEEauthorblockN{Souradip Chakraborty\textsuperscript{\textsection}}
\IEEEauthorblockA{Walmart Labs\\
souradip24@gmail.com\\
}
\and
\IEEEauthorblockN{Aritra Roy Gosthipaty\textsuperscript{\textsection}}
\IEEEauthorblockA{Netaji Subhash Engineering College\\
aritrarg.ece2016@nsec.ac.in}
\and
\IEEEauthorblockN{Sayak Paul\textsuperscript{\textsection}}
\IEEEauthorblockA{PyImageSearch\\
spsayakpaul@gmail.com}
}

\maketitle
\begingroup\renewcommand\thefootnote{\textsection}
\footnotetext{Equal contribution}
\endgroup

\maketitle
\begin{abstract}
In the realms of computer vision,  it is evident that deep neural networks perform better in a supervised setting with a large amount of labeled data. The representations learned with supervision are not only of high quality but also helps the model in enhancing its accuracy. However, the collection and annotation of a large dataset are costly and time-consuming. To avoid the same, there has been a lot of research going on in the field of unsupervised visual representation learning especially in a self-supervised setting. Amongst the recent advancements in self-supervised methods for visual recognition, in SimCLR Chen et al. shows that good quality representations can indeed be learned without explicit supervision. In SimCLR, the authors maximize the similarity of augmentations of the same image and minimize the similarity of augmentations of different images. A linear classifier trained with the representations learned using this approach yields 76.5\% top-1 accuracy on the ImageNet ILSVRC-2012 dataset. In this work, we propose that, with the normalized temperature-scaled cross-entropy (\textit{NT-Xent}) loss function (as used in SimCLR), it is beneficial to not have images of the same category in the same batch. In an unsupervised setting, the information of images pertaining to the same category is missing. We use the latent space representation of a denoising autoencoder trained on the unlabeled dataset and cluster them with \(k\)-means to obtain pseudo labels. With this apriori information we batch images, where no two images from the same category are to be found. We report comparable performance enhancements on the CIFAR10 dataset and a subset of the ImageNet dataset \footnote{\href{https://github.com/thunderInfy/imagenet-5-categories}{https://github.com/thunderInfy/imagenet-5-categories}}. We refer to our method as \textit{G-SimCLR} \footnote{Code available at \href{https://github.com/ariG23498/G-SimCLR}{\textcolor{mypink1}{https://github.com/ariG23498/G-SimCLR}}} \footnote{This paper is accepted at ICDM 2020 DLKT workshop.}. 
\end{abstract}


\IEEEpeerreviewmaketitle

\section{Introduction}
The huge success of deep learning algorithms in computer vision is heavily dependent on the quality and the information content of the latent visual representations learned from the input images. There has been almost a decade of very successful research in the field of supervised representation learning from images using different variants of convolutional neural networks \cite{NIPS2012_4824, simonyan2014deep, szegedy2015rethinking, he2015deep, tan2019efficientnet}. However, the complex parametric structure of deep convolutional neural networks makes it heavily dependent on the amount of labeled training data. Though, humongous amount of labeled data helps in increasing both the accuracy of these models and the information content of the visual representation learned in these models, the collection and labeling of a large dataset are extremely expensive and time-consuming. However, the amount of unlabeled data (e.g. images on the Internet) is practically unlimited in nature except for some domains like healthcare. Hence to remedy the expensive efforts of creating a huge amount of labeled datasets, there has been an increasing amount of research dedicated to the field of unsupervised (i.e. where no explicit label information is used) visual representation learning. Methods that have come out as a byproduct of this research effort,  have shown tremendous performance with respect to their fully supervised counterparts. Features learned by these methods have been very useful for several downstream visual recognition tasks such as image classification, object detection, instance segmentation, and so on even with very less labeled data \cite{trinh2019selfie, misra2019selfsupervised, he2019momentum, chen2020simple, grill2020bootstrap}. 

When dealing with unsupervised regimes, methods that are based on self-supervised learning have shown great promise. In self-supervised learning, explicit labels are not used; instead, we utilize the unlabeled dataset to formulate an auxiliary supervised learning problem e.g. predicting rotation angle from images, placing image patches in the right place, placing video frames in the right order, and so on \cite{gidaris2018unsupervised, noroozi2016unsupervised, lee2017unsupervised}. 

Another approach that we see in SimCLR \cite{chen2020simple} is contrasting different views of images and training models to maximize the agreement between the views of the same images while minimizing the agreement between the views of other images. A linear classifier trained with the representations learned using this approach yields a 76.5\% top-1 accuracy on the ImageNet ILSVRC-2012 dataset \cite{ILSVRC15}.

However, in \cite{khosla2020supervised} Khosla et al. show that SimCLR being dependent on the sampling probability might learn representations that can distant images from the same label space if they appear in the same training batch. This in turn hurts the representations learned for subsequent downstream tasks. This can turn out to be a crucial problem if the downstream task is highly imbalanced in its label space. For example, in a dataset with 5 classes if one class is a majority class having the maximum number of training samples, then there is a very high probability that samples from that majority class will be in the same batch while training SimCLR. If images belonging to the same class are sent to the same batch for training in SimCLR, the model might learn representations that can distant the images of the same class in the latent space which can hurt the quality of the learned representations. 

Motivated by these findings and work present in \cite{caron2018deep} by Caron et al. we propose a simplistic approach for generating prior information of the label space without using any explicit labels and then trying to enhance the quality of the representations learned via SimCLR through this apriori information. We use a denoising autoencoder (fully convolutional) \cite{Vincent2008} in a self-supervised setting where we minimize the reconstruction loss between the input and noisy images and extract the latent space representations from the trained autoencoder. We leverage this latent feature map from the autoencoder to cluster the input space of images and assign each input image to the corresponding cluster which we refer to as pseudo labels. We incorporate this prior pseudo label information while determining the training batches for  SimCLR which minimizes the risk of getting images from the same label space in the same batch. We report comparable performance gains achieved using G-SimCLR i.e. the proposed method on two datasets - CIFAR10 \cite{learning-features-2009-TR} and a subset of ImageNet \cite{imagenet-subset} (referred to as ImageNet Subset throughout the paper). To encourage research along similar lines we also open-source the code to reproduce our experiments.

\section{Methodology}
In SimCLR, Chen et al. have shown that a strong suite of stochastic data augmentation operations, suitable non-linear transformations and contrastive learning with large batch size, and longer training time enhance the quality of the learned representations significantly. In this work, we present an additional step on top of the SimCLR framework to ensure two semantically similar images do not get treated differently. The schematics of this additional step is presented below - 
\begin{itemize}
    \item First, we train a denoising autoencoder (fully convolutional) on the given dataset.
    \item In the second step, we take the encoder-learned robust representations (also referred to as latent space representations) and use \(k\)-means to get initial cluster assignments of those representations. We refer to these cluster assignments as pseudo labels for the given dataset.
    \item Finally, we use these pseudo labels to prepare the batches and use them for SimCLR training.
\end{itemize}

\begin{figure}[h]
\includegraphics[width=8cm]{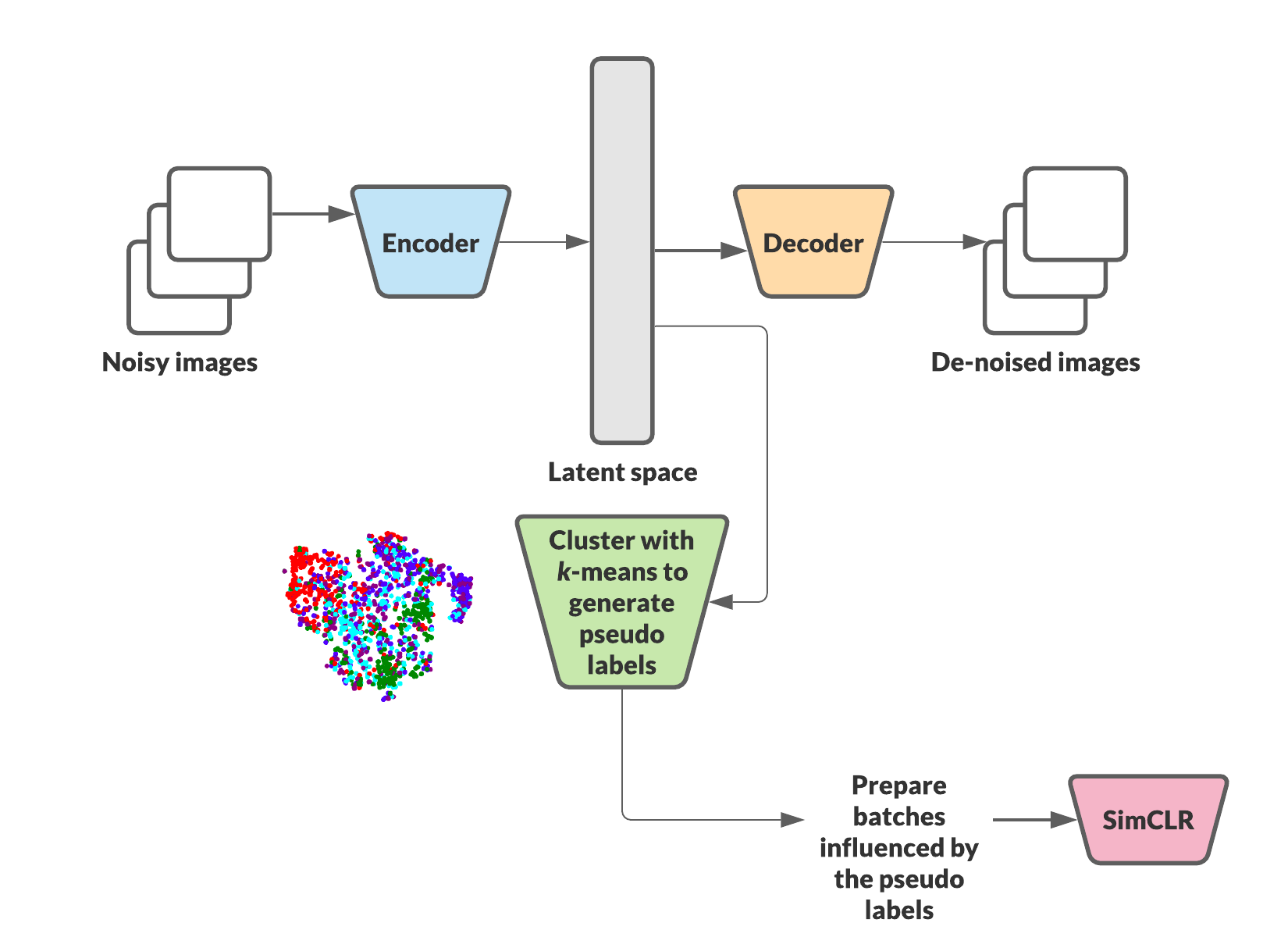}
\caption{The proposed method involves training a denoising autoencoder to capture a latent space representation for the given dataset. We then cluster this latent space to generate pseudo labels for the given dataset. Finally, we use these pseudo labels to prepare the data batches for SimCLR training on the given dataset.}
\label{fig:methodology}
\end{figure}

We present this methodology in Figure \ref{fig:methodology}. In the first step of our methodology, a denoising autoencoder is used which enhances the robustness of the latent feature representation when compared to a vanilla autoencoder. This is done by distorting the input image(x) to get a semi-distorted version of the same by means of Gaussian noise. The corrupted input($\hat{x}$) is then mapped, as in case of a vanilla autoencoder, to a latent representation $z = f(\hat{x})$, where $z \in R^{d}$. The latent encoder representation $z$ now helps in reconstructing the distortion-free image input in the decoding phase and the average reconstruction loss is minimized to learn the parameters of the denoising autoencoder. 

Once the denoising autoencoder is trained, the encoder representations are extracted from all the input images. This representation of the input data space is denoted by $Y \in R^{n\times d}$, where $n$ is the total number of input images and $d$ is the dimensionality of the encoder representation. We use \(k\)-means on this representation space of the images ($Y$) to get dense clusters (assume the number of clusters to be $k$). The clusters formed by minimizing the within-group variance and maximizing the between-group variance, ensures that the images which are similar to each other fall in the same cluster and which are different falls in different clusters. The cluster assignment is given by $d_{ij} = 1$, when $i_{th}$ image falls in the $j_{th}$ cluster else $d_{ij} = 0$. These assignments are considered as pseudo labels in our work, used for selecting the images for each batch in training SimCLR. 

In the final step of our methodology, instead of randomly sampling the mini-batch of $p$ samples, we use the pseudo labels (obtained by \(k\)-means clustering on the autoencoder representations) to sample the images in each of the batches. In order to minimize the risk of having similar images in the same batch while SimCLR training, which might adversely impact the representations, we use a pseudo label based stratified sampling methodology in preparing the training batches. In our current implementation, we have kept the batch size and cluster number to be the same, $k$. So, while preparing the training batches, we perform random sampling (without replacement), stratified by the pseudo labels. \(k\)-means clustering does not ensure equal-sized clusters. This means that some batches will have more from one cluster and less from others. This ensures that our method does not provide a hard constraint on the images of the batches to belong to different and discrete classes only. 

We then follow the same training objective as SimCLR but on the pairs of augmented examples derived from the batches formed using the aforementioned methodology. The loss function (\textit{NT-XEnt}) for training SimCLR (as well as G-SimCLR) is expressed as

\begin{equation}
\ell _{i,j} =-\log\frac{\exp\left(\operatorname{sim}(\boldsymbol{z}_{i} ,\boldsymbol{z}_{j}) /\tau \right)}{\sum ^{2N}_{k=1}\mathbbm{1}_{[ k\neq i]}\exp\left(\operatorname{sim}(\boldsymbol{z}_{i} ,\boldsymbol{z}_{k}) /\tau \right)}
\end{equation}

where $sim$ denotes cosine similarity, $z$ denotes the outputs from the non-linear projection head as used in the original SimCLR work, and $\tau$ denotes the temperature hyperparameter. 

Overall, the primary additional step as discussed above does introduce an additional bottleneck in terms of computational complexity and execution time. We do not put too much focus on generating the pseudo labels, these are used only to guide the batch formation systematically. In the original SimCLR work large batch sizes (typically 4096) are used to have enough negative samples. However, maintaining such large batch sizes is computationally expensive and G-SimCLR provides a simple way to remedy that.

\section{Experimental setup}
\textbf{\textit{Denoising autoencoder.}} A Gaussian noise centered at 0 with a standard deviation of 0.01 is provided as the input layer of the denoising autoencoder. This layer adds Gaussian noise to the input images. The autoencoder learns from the mean squared error loss (\textit{MSE} loss) between the noisy and the original images. Adam optimizer \cite{kingma2014adam} is used to update the weights of the network. Using early stopping (with a patience of 5) the autoencoder trains for 100 epochs. After the autoencoder is trained, the latent space representation of each training image is extracted. The representations are taken from the bottleneck of the autoencoder.

\textbf{\textit{Clustering with \(k\)-means.}} By applying \(k\)-means clustering algorithm on the latent space representations of images, we retrieve the cluster labels for each image. The choice for the number of clusters is 64. Empirically, we found out that it is helpful to have a cluster number that is large enough to capture the implicit marginal distribution of the given dataset. One could also use domain knowledge to determine this value \cite{asano2019selflabelling}. More rigorous experimentation to study the effect of the number of \(k\) on the contrastive learning task and further downstream tasks has been left as a scope for future work. 

\textbf{\textit{SimCLR and G-SimCLR.}} Our approach of SimCLR training is not exactly similar to what is presented in the original paper. We refer to this variant of SimCLR as \textit{SimCLR with minor modifications}. ResNet \cite{he2015deep} is used as the feature backbone network to extract meaningful representations from the input images (and their augmentation variants). Similar to SimCLR we map these representations to a lower-dimensional space using a shallow fully-connected network with ReLU non-linearity. We use a subset of augmentation operations used in SimCLR. Specifically, we use random horizontal flips and color distortions. The optimizer used is stochastic gradient descent with a cosine learning rate decay \cite{loshchilov2016sgdr} instead of LARS \cite{you2017large}. A temperature ($\tau$) value of 0.1 is used. 

\textbf{\textit{Linear evaluation and label efficient fine-tuning.}} For linear evaluation of the representations learned using our variant of SimCLR and G-SimCLR we only use a linear layer. We further fine-tune the representations with \textit{10\% labeled examples} from the datasets we use. We report the final performances from these experiments in Tables \ref{tab:linear-eval} and \ref{tab:fine-tune} respectively. For these experiments, we use early stopping and train with the default values of the Adam optimizer in Keras \footnote{\href{https://keras.io/api/optimizers/adam/}{https://keras.io/api/optimizers/adam/}}. 

Our variant of SimCLR and G-SimCLR differ only in the way the training batches are made. In our proposed approach, the training images are ordered in a batch of 64 samples. The organization of images in a batch is such that no two images belonging to the same cluster are to be found in the same batch. The choice of 64 as the batch size is not cherry-picked. We use 64 clusters so that in a batch of 64 images, we can group images belonging to distinct clusters in a batch.

Below we present dataset-specific configuration details. 

\subsection{CIFAR10}
The latent space of the autoencoder is (4,4,128) dimensional. Our variant of SimCLR and G-SimCLR are trained for 15 epochs. For the feature backbone network, we use a variant of ResNet that is specifically suited for the CIFAR10 dataset \footnote{We use this implementation - \href{https://github.com/GoogleCloudPlatform/keras-idiomatic-programmer/blob/master/zoo/resnet/resnet_cifar10.py}{https://github.com/GoogleCloudPlatform/keras-idiomatic-programmer/blob/master/zoo/resnet/resnet\_cifar10.py}}.

\subsection{ImageNet Subset}
We experiment with two autoencoders for this dataset. With the first kind, we use 3 hidden layers for encoder and decoder respectively. The latent space here is (28,28,128) dimensional. The second kind has 5 hidden layers in the encoder and the decoder. The latent space here is (7,7,1024) dimensional. We find that the deeper autoencoder encapsulates more robust and discriminative visual representations in its latent space. We use ResNet50 \footnote{We use \texttt{tf.keras.applications.ResNet50} as seen here - \href{https://keras.io/api/applications/resnet/}{https://keras.io/api/applications/resnet/}.} as the feature backbone network and we train it for 200 epochs. 

\section{Experimental results}
\begin{figure}[h]
\includegraphics[width=8cm]{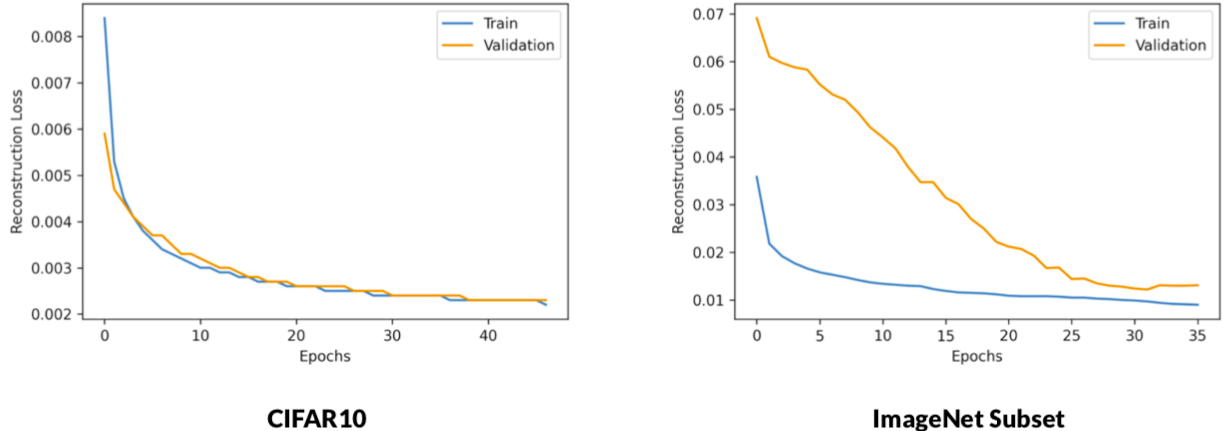}
\caption{Train and validation reconstruction losses of the denoising autoencoder networks trained on the CIFAR10 and ImageNet Subset datasets respetively.}
\label{fig:AE_Loss}
\end{figure}

We observe the performance of the fully convolutional denoising autoencoder on a given dataset in terms of its reconstruction loss. Since our approach is completely unsupervised in nature, the train and validation reconstruction losses give us a directional understanding of the quality of the representations learned with the autoencoder network as shown in \ref{fig:AE_Loss}. The convergence of the train and validation reconstruction losses indicate that the representations learned are meaningful and informative.

\begin{figure}[h]
\includegraphics[width=8cm]{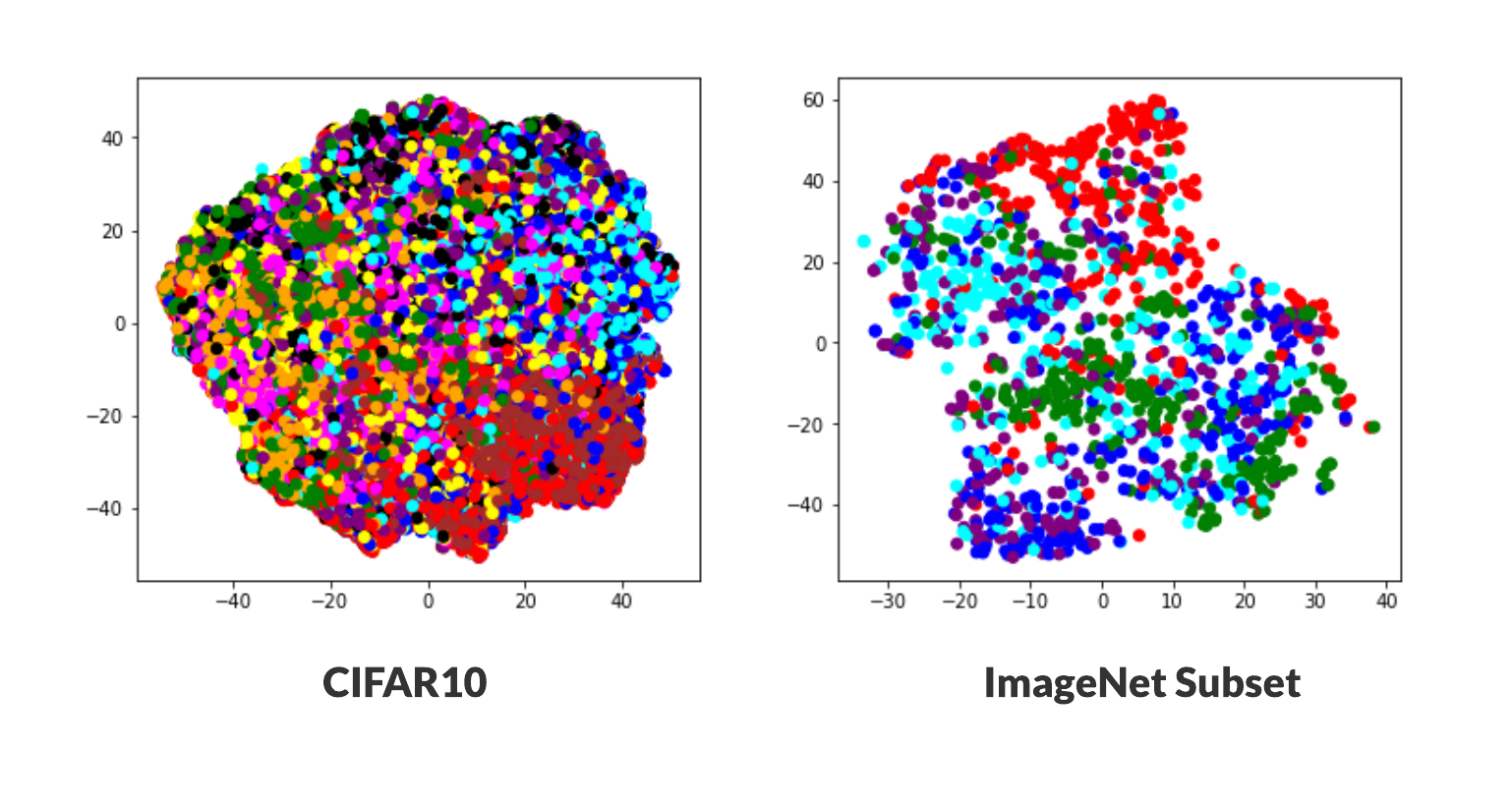}
\caption{t-SNE projections of the latent space representations learned by the denoising autoencoder on the CIFAR10 and ImageNet Subset datasets.}
\label{fig:tsne-projections}
\end{figure}

In an unsupervised setting, we do not know how many categories the images belong to. In our approach, we cluster the latent space representations of images into 64 clusters. The image representations close to each other fall into the same cluster, while the ones that are far away, belong to different clusters. From the t-SNE \cite{vanDerMaaten2008} projections (as shown in Figure \ref{fig:tsne-projections}) of the latent space representations (as learned by the denoising autoencoder), it is evident that the representations of similar images cluster together. We do not make use of the typical cluster evaluation metrics like Normalized Mutual Information Score \cite{nmi} and Rand Index \cite{rand} since they involve comparisons with the ground-truth labels, thereby eliminating the need for supervision as much as possible. We do not treat the clustering problem very rigorously; we make use of it just enough to generate the pseudo labels needed to systematically construct the mini-batches for SimCLR. 

In Table \ref{tab:cluster-labels} we show how we represent the pseudo labels for G-SimCLR.

\begin{table}[!t]
\centering
\caption{Image indices and their cluster labels as generated by \(k\)-means. We treat these cluster labels as pseudo labels in G-SimCLR.}
\label{tab:cluster-labels}
\begin{tabular}{|c|c|}
\hline
\textbf{Image index} & \textbf{Cluster label} \\ \hline
36 & 0 \\ \hline
61 & 15 \\ \hline
99 & 47 \\ \hline
11 & 48 \\ \hline
47 & 14 \\ \hline
\end{tabular}
\end{table}

\begin{figure}[h]
\includegraphics[width=9cm]{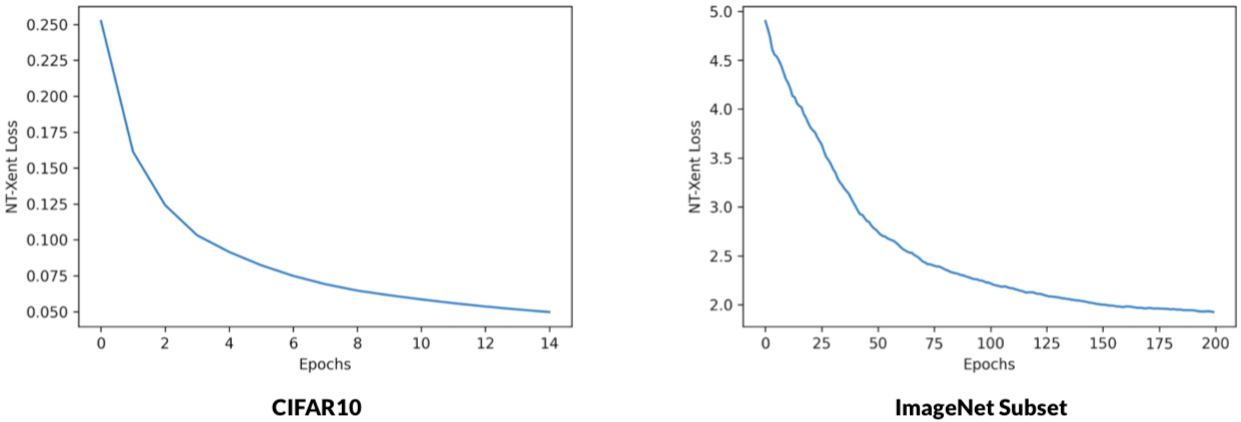}
\caption{Loss (\textit{NT-Xent}) curves as obtained from the G-SimCLR training with the CIFAR10 and ImageNet Subset datasets respectively.}
\label{fig:loss-curves}
\end{figure}

Figure \ref{fig:loss-curves} shows the loss (\textit{NT-Xent}) curves as obtained from the G-SimCLR training with the CIFAR10 and ImageNet Subset datasets. We can see that in both cases the loss decrease is steady suggesting good progress in G-SimCLR training.

To compare G-SimCLR with our variant of SimCLR, we run two downstream supervised classification tasks on the visual representations that each of them learns. We first run linear evaluation on the classification task and later fine-tune the models with 10 \% data and then evaluate again.

For linear evaluation we report the validation accuracy in Table \ref{tab:linear-eval} for different levels of the feature backbone network and the projection head denoted as \textbf{P1}, \textbf{P2} and \textbf{P3}  where 
\begin{itemize}
    \item \textbf{P1} denotes the feature backbone network + the entire non-linear projection head - its final layer
    \item \textbf{P2} denotes the feature backbone network + the entire non-linear projection head - its final two layers
    \item \textbf{P3} denotes the feature backbone network only
\end{itemize}

Next in Table \ref{tab:fine-tune} we report the validation accuracy where the representations learned by our variant of SimCLR and G-SimCLR are fine-tuned with only 10\% labeled training examples. In this case, we do not use any layer from the non-linear projection head. The results reported suggest that the intuition behind G-SimCLR is valid.

The results reported suggest that the guided apriori provided to batches in G-SimCLR is indeed more performant than our SimCLR variant.

\begin{table}[!t]
\caption{Performance of the linear classifiers trained on top of the representations (kept frozen during training linear classifiers) learned by G-SimCLR.}
\label{tab:linear-eval}
\centering
\begin{tabular}{@{}|c|c|c|c|@{}}
\toprule
\multicolumn{4}{|c|}{Linear evaluation} \\ \midrule
 &  & CIFAR10 & ImageNet Subset \\ \midrule
Fully supervised &  & 73.62 & 67.6 \\ \midrule
\multirow{3}{*}{SimCLR with minor modifications} & \multicolumn{1}{l|}{P1} & 37.69 & 52.8 \\ \cmidrule(l){2-4} 
 & \multicolumn{1}{l|}{P2} & 39.4 & 48.4 \\ \cmidrule(l){2-4} 
 & \multicolumn{1}{l|}{P3} & 39.92 & 52.4 \\ \midrule
\multirow{3}{*}{G-SimCLR (ours)} & P1 & \textbf{38.15} & \textbf{56.4} \\ \cmidrule(l){2-4} 
 & P2 & \textbf{41.01} & \textbf{56.8} \\ \cmidrule(l){2-4} 
 & P3 & \textbf{40.5} & \textbf{60.0} \\ \bottomrule
\end{tabular}
\end{table}

\begin{table}[!t]
\caption{Performance of weakly supervised classifiers trained on top of the representations learned by G-SimCLR. We fine-tuned the representations with only 10\% labeled data.}
\label{tab:fine-tune}
\centering
\begin{tabular}{@{}|c|c|c|@{}}
\toprule
\multicolumn{3}{|c|}{Fine-tuning (10\% labeled data)} \\ \midrule
 & CIFAR 10 & ImageNet Subset \\ \midrule
SimCLR with minor modifications & 42.21 & 49.2 \\ \midrule
G-SimCLR (ours) & \textbf{43.1} & \textbf{56.0} \\ \bottomrule
\end{tabular}
\end{table}

\section{Conclusion}
In this work, we build on top of the original SimCLR work and present a simple yet effective way to form robust mini-batches for SimCLR training. We study the main components used in G-SimCLR (proposed method) i.e. denoising autoencoder and the intermediary clustering step and we report comparable performance enhancements on two datasets.  

We are hopeful G-SimCLR will be able to remedy the problems of SimCLR e.g. requirement of maintaining large batch sizes, false positives in the learned representation space, and so on. Benchmarking G-SimCLR on more standardized datasets (such as ImageNet, Flowers, Caltech-101, etc.), more detailed study on the autoencoder-learned latent space representations, using more efficient algorithms for clustering remain as future scopes of improvements for us.


\bibliography{bibliography} 

\begin{thebibliography}{10}

\bibitem{NIPS2012_4824}
A.~Krizhevsky, I.~Sutskever, and G.~E. Hinton, ``Imagenet classification with
  deep convolutional neural networks,'' in {\em Advances in Neural Information
  Processing Systems 25} (F.~Pereira, C.~J.~C. Burges, L.~Bottou, and K.~Q.
  Weinberger, eds.), pp.~1097--1105, Curran Associates, Inc., 2012.

\bibitem{simonyan2014deep}
K.~Simonyan and A.~Zisserman, ``Very deep convolutional networks for
  large-scale image recognition,'' 2014.

\bibitem{szegedy2015rethinking}
C.~Szegedy, V.~Vanhoucke, S.~Ioffe, J.~Shlens, and Z.~Wojna, ``Rethinking the
  inception architecture for computer vision,'' 2015.

\bibitem{he2015deep}
K.~He, X.~Zhang, S.~Ren, and J.~Sun, ``Deep residual learning for image
  recognition,'' 2015.

\bibitem{tan2019efficientnet}
M.~Tan and Q.~V. Le, ``Efficientnet: Rethinking model scaling for convolutional
  neural networks,'' 2019.

\bibitem{trinh2019selfie}
T.~H. Trinh, M.-T. Luong, and Q.~V. Le, ``Selfie: Self-supervised pretraining
  for image embedding,'' 2019.

\bibitem{misra2019selfsupervised}
I.~Misra and L.~van~der Maaten, ``Self-supervised learning of pretext-invariant
  representations,'' 2019.

\bibitem{he2019momentum}
K.~He, H.~Fan, Y.~Wu, S.~Xie, and R.~Girshick, ``Momentum contrast for
  unsupervised visual representation learning,'' 2019.

\bibitem{chen2020simple}
T.~Chen, S.~Kornblith, M.~Norouzi, and G.~Hinton, ``A simple framework for
  contrastive learning of visual representations,'' 2020.

\bibitem{grill2020bootstrap}
J.-B. Grill, F.~Strub, F.~Altché, C.~Tallec, P.~H. Richemond, E.~Buchatskaya,
  C.~Doersch, B.~A. Pires, Z.~D. Guo, M.~G. Azar, B.~Piot, K.~Kavukcuoglu,
  R.~Munos, and M.~Valko, ``Bootstrap your own latent: A new approach to
  self-supervised learning,'' 2020.

\bibitem{gidaris2018unsupervised}
S.~Gidaris, P.~Singh, and N.~Komodakis, ``Unsupervised representation learning
  by predicting image rotations,'' 2018.

\bibitem{noroozi2016unsupervised}
M.~Noroozi and P.~Favaro, ``Unsupervised learning of visual representations by
  solving jigsaw puzzles,'' 2016.

\bibitem{lee2017unsupervised}
H.-Y. Lee, J.-B. Huang, M.~Singh, and M.-H. Yang, ``Unsupervised representation
  learning by sorting sequences,'' 2017.

\bibitem{ILSVRC15}
O.~Russakovsky, J.~Deng, H.~Su, J.~Krause, S.~Satheesh, S.~Ma, Z.~Huang,
  A.~Karpathy, A.~Khosla, M.~Bernstein, A.~C. Berg, and L.~Fei-Fei, ``{ImageNet
  Large Scale Visual Recognition Challenge},'' {\em International Journal of
  Computer Vision (IJCV)}, vol.~115, no.~3, pp.~211--252, 2015.

\bibitem{khosla2020supervised}
P.~Khosla, P.~Teterwak, C.~Wang, A.~Sarna, Y.~Tian, P.~Isola, A.~Maschinot,
  C.~Liu, and D.~Krishnan, ``Supervised contrastive learning,'' 2020.

\bibitem{caron2018deep}
M.~Caron, P.~Bojanowski, A.~Joulin, and M.~Douze, ``Deep clustering for
  unsupervised learning of visual features,'' 2018.

\bibitem{Vincent2008}
P.~Vincent, H.~Larochelle, Y.~Bengio, and P.-A. Manzagol, ``Extracting and
  composing robust features with denoising autoencoders,'' in {\em
  International Conference on Machine Learning proceedings}, 2008.

\bibitem{learning-features-2009-TR}
A.~Krizhevsky, ``Learning multiple layers of features from tiny images.''
  \url{https://www.cs.toronto.edu/~kriz/learning-features-2009-TR.pdf}, 2009.

\bibitem{imagenet-subset}
A.~Rastogi, ``Imagenet 5 categories.''
  \url{https://github.com/thunderInfy/imagenet-5-categories}, 2020.

\bibitem{kingma2014adam}
D.~P. Kingma and J.~Ba, ``Adam: A method for stochastic optimization,'' 2014.

\bibitem{asano2019selflabelling}
Y.~M. Asano, C.~Rupprecht, and A.~Vedaldi, ``Self-labelling via simultaneous
  clustering and representation learning,'' 2019.

\bibitem{loshchilov2016sgdr}
I.~Loshchilov and F.~Hutter, ``Sgdr: Stochastic gradient descent with warm
  restarts,'' 2016.

\bibitem{you2017large}
Y.~You, I.~Gitman, and B.~Ginsburg, ``Large batch training of convolutional
  networks,'' 2017.

\bibitem{vanDerMaaten2008}
L.~van~der Maaten and G.~Hinton, ``Visualizing data using {t-SNE},'' {\em
  Journal of Machine Learning Research}, vol.~9, pp.~2579--2605, 2008.

\bibitem{nmi}
A.~Strehl and J.~Ghosh, ``Cluster ensembles --- a knowledge reuse framework for
  combining multiple partitions,'' {\em JMLR}, 2003.

\bibitem{rand}
W.~Rand, ``Objective criteria for the evaluation of clustering methods. journal
  of the american statistical association.,'' {\em JASA}, 1971.

\end{thebibliography}
\bibliographystyle{ieeetr}

\end{document}